%% file: main.tex
\documentclass[sigconf,dvipsnames]{acmart}

\AtBeginDocument{%
  \providecommand\BibTeX{{%
    \normalfont B\kern-0.5em{\scshape i\kern-0.25em b}\kern-0.8em\TeX}}}
\usepackage{wrapfig, booktabs}
\copyrightyear{2022}
\acmYear{2022}
\setcopyright{rightsretained}
\acmConference[FAccT '22]{2022 ACM Conference on Fairness, Accountability, and Transparency}{June 21--24, 2022}{Seoul, Republic of Korea}
\acmBooktitle{2022 ACM Conference on Fairness, Accountability, and Transparency (FAccT '22), June 21--24, 2022, Seoul, Republic of Korea}\acmDOI{10.1145/3531146.3533148}
\acmISBN{978-1-4503-9352-2/22/06}

\usepackage{xcolor}

\begin{document}


\title[Learning to Limit Data Collection via Scaling Laws]{Learning to Limit Data Collection via Scaling Laws: \linebreak A Computational Interpretation for the Legal Principle of \linebreak Data Minimization}

\author{Divya Shanmugam}
\email{divyas@mit.edu}
\affiliation{%
  \institution{Massachusetts Institute of Technology}
  \city{Cambridge}
  \state{MA}
  \country{USA}
  \postcode{02139}
}

\author{Samira Shabanian}
\email{samira.shabanian@microsoft.com}
\affiliation{%
  \institution{Microsoft Research}
  \city{Montreal}
  \country{Canada}
}

\author{Fernando Diaz}
\authornote{Now at Google.}
\email{diazf@acm.org}
\affiliation{%
  \institution{Microsoft Research}
  \city{Montreal}
  \country{Canada}
}

\author{Michèle Finck}
\email{m.finck@uni-tuebingen.de}
\affiliation{%
  \institution{University of Tübingen}
  \country{Germany}
}

\author{Asia J. Biega}
\email{asia.biega@mpi-sp.org}
\affiliation{%
  \institution{Max Planck Institute for Security and Privacy}
  \city{Bochum}
  \country{Germany}
}


\begin{abstract}


Modern machine learning systems are increasingly characterized by extensive personal data collection, despite the diminishing returns and increasing societal costs of such practices. Yet, data minimisation is one of the core data protection 
principles enshrined in the European Union's General Data Protection 
Regulation ('GDPR') and requires that only personal data that is 
adequate, relevant and limited to what is necessary is processed.  However, the principle has seen limited adoption due to the lack of technical interpretation. 
 
 In this work, we build on literature in machine learning and law to propose \textbf{FIDO}, a \textbf{F}ramework for \textbf{I}nhibiting \textbf{D}ata \textbf{O}vercollection. FIDO learns to limit data collection based on an interpretation of data minimization tied to system performance. Concretely, FIDO provides a data collection stopping criterion by iteratively updating an estimate of the \emph{performance curve}, or the relationship between dataset size and performance, as data is acquired. FIDO estimates the performance curve via a piecewise power law technique that models distinct phases of an algorithm's performance throughout data collection \emph{separately}. Empirical experiments show that the framework produces accurate performance curves and data collection stopping criteria across datasets and feature acquisition algorithms. We further demonstrate that many other families of curves systematically \emph{overestimate} the return on additional data. Results and analysis from our investigation offer deeper insights into the relevant considerations when designing a data minimization framework, including the impacts of active feature acquisition on individual users and the feasability of user-specific data minimization. We conclude with practical recommendations for the implementation of data minimization. 

\end{abstract}


\maketitle

\input{99_notation.tex}

\input{01_introduction}

\input{02_1_background_legal}
\input{09_related_work}

\input{02_2_problem_statement}
\input{04_method_three_stage}
\input{05_experiment_setup}

\input{06_experiments}
\input{08_pitfalls}
\input{10_putting_it_all_together}
\input{11_limitations}

\input{12_conclusion_old}


\begin{acks}
We thank John Guttag, Hansa Srinivasan, and Hal Daumè III for helpful comments.
\end{acks}

\bibliographystyle{ACM-Reference-Format}
\bibliography{main}


\end{document}

%% file: 99_notation.tex
\newcommand{\vagueref}[1]{{the supplementary material}}

\newcommand{\dmSet}[1]{\mathcal{#1}}
\newcommand{\dmSetSize}[1]{|\dmSet{#1}|}
\newcommand{\dmUnion}[2]{#1 \cup #2}
\newcommand{\dmEstimate}[1]{\hat{#1}}

\newcommand{\dmPool}[0]{\dmSet{P}}
\newcommand{\dmTraining}[0]{\dmSet{I}}
\newcommand{\dmValidation}[0]{\dmSet{V}}
\newcommand{\dmHistoricData}[0]{\dmSet{A}}

\newcommand{\dmUsers}[0]{\dmSet{U}}
\newcommand{\dmUser}[0]{u}

\newcommand{\dmModel}[0]{M}

\newcommand{\dmBoundFn}[0]{f}
\newcommand{\dmLowerBound}[0]{\dmBoundFn_{0}}
\newcommand{\dmMiddleBound}[0]{\dmBoundFn_{1}}
\newcommand{\dmUpperBound}[0]{\dmBoundFn_{2}}

\newcommand{\dmBoundParameterA}[0]{A}
\newcommand{\dmBoundParameterB}[0]{b}

\newcommand{\dmReturn}[0]{t}

\newcommand{\dmPerformanceFn}[0]{\sigma_\dmModel}
\newcommand{\dmPerformance}[1][]{\dmPerformanceFn}

\newcommand{\dmPerformanceTreatment}[0]{\dmPerformance[\dmUnion{\dmTraining}{\dmPolicy(\dmPool,\dmBudget)}]}

\newcommand{\dmUserPerformanceFn}[0]{\sigma_{\dmModel}^{\dmUser}}
\newcommand{\dmUserPerformance}[1][]{\dmUserPerformanceFn}

\newcommand{\dmPerformanceStarting}[0]{\dmPerformance[\dmTraining]}
\newcommand{\dmPerformanceTarget}[0]{\dmPerformance[\dmUnion{\dmTraining}{\dmPool}]}

\newcommand{\dmNumQueries}[0]{q}

\newcommand{\dmBudget}[0]{n}

\newcommand{\dmGainThreshold}[0]{g}
\newcommand{\dmGain}[0]{p}
\newcommand{\dmGainEstimate}[0]{\dmEstimate{\dmGain}}

\newcommand{\dmSlopeThreshold}[0]{t}
\newcommand{\dmSlope}[0]{s}
\newcommand{\dmSlopeEstimate}[0]{\dmEstimate{\dmSlope}}

\newcommand{\dmPolicy}[0]{H}

%% file: 01_introduction.tex
\section{Introduction}
\label{sec:introduction}

Data minimisation is a core principle of the European Union's Data Protection Regulation~\citep{regulation2016regulation}, as well as data protection laws in other jurisdictions: 
\begin{quote}
"Personal data shall be: [...] adequate, relevant and limited to what is necessary in relation to the purposes for which they are processed (data minimisation)"
\end{quote}
%




The requirement serves as a guideline for respectfully processing data. Recent empirical research has shown that it is possible to replicate the performance of data-driven systems with significantly less data~\cite{biega2020dm, chow2013differential,vincent2019data, wen2018exploring}. These findings are a consequence of the diminishing returns that data collection exhibits across applications and domains~\cite{hestness1712deep, krause2010utility,tae2020slice}. Recognizing that limiting data is possible, legal guidelines point to algorithmic techniques that could be incorporated into minimization pipelines, including feature selection~\cite{icoDataMinimizationTechniques} or examination of learning curves~\cite{norwegianAuthorityArtificial}.

Yet, despite the existence of numerous algorithmic techniques that could be adapted to comply with the minimization requirement, the data minimisation principle has received little attention from the computer science community to date. As noted by scholars reviving discussion about the principle, a dearth of concrete mathematical definitions and guidelines is one of the main factors inhibiting adoption~\cite{finckreviving}.
Indeed, qualitative research has shown a lack of consistent data minimization standards or an understanding of the principle among software developers~\cite{senarath2018understanding}.

Recent interpretations of data minimization propose to tie the data collection purpose in data-driven systems to performance metrics, an interpretation termed \emph{performance-based data minimization}~\cite{finckreviving,biega2020dm}. Our work follows this interpretation, addressing the question of \emph{how to proactively satisfy the performance-based data minimization principle in machine learning with personal data}. \\

\textbf{Contributions.} In this work, we propose FIDO, a Framework for Inhibiting Data Overcollection, and demonstrate how ongoing personal data collection could be approached in the context of data minimization. 

FIDO's key conceptual proposal is to \emph{adaptively learn an algorithm's performance curve} so that an \emph{appropriate data collection stopping point} can be determined accurately. By modeling the performance curve directly, the framework remains flexible to different underlying feature acquisition algorithms and definitions of performance. Experiments involving multiple datasets in the recommender systems domain validate this flexibility and demonstrate FIDO's ability to estimate algorithm performance given additional data accurately \emph{without} collecting more data.

The core technical insight FIDO contributes lies in its performance curve estimation procedure, which builds on recent work in machine learning literature that identifies three distinct phases in the performance curves of learning algorithms: the small data phase, the power law phase, and the diminishing returns phase \cite{hestness1712deep}. We demonstrate that these performance phases can be observed in the context of user data collection and provide a technique to model the data collection phases directly by adaptively learning a piecewise power law curve. Empirical experiments show that other approaches to estimate performance curves systematically \emph{overestimate} the return on additional data. Modeling each phase directly allows FIDO to not only learn the performance curve more accurately as data is acquired but also satisfy practical constraints of data minimization.


Finally, we examine issues related to user-level data minimization. We demonstrate the impacts that algorithmic components such as active feature acquisition (a technique to intelligently select which feature values to acquire)  might have on minimization outcomes. We find that active feature acquisition can lead to unequal, concentrated data collection from a small set of users, in addition to decreased minimization performance for evolving user communities or when sensitive features are excluded from initial data collection. We also find that user-specific performance curves are highly variable, where the collection of more data can often result in a \emph{decrease} in user-specific performance. 

This paper seeks to offer a computational perspective on the GDPR's principle of data minimization, contributing insights to the ongoing discussions about the technical implementation of this core data protection principle, and to provide recommendations to practitioners and scholars moving forward.

%% file: 02_1_background_legal.tex
\section{Legal Background}

Data minimization is one of the core principles of European data protection law. In recent years, many have questioned its suitability in face of technical advances based on the repurposing of large quantities of data\cite{koops4trouble}. Some indeed fear that adherence to the principle ‘would sacrifice considerable social benefit’ as it may limit the innovative potential of machine learning~\cite{maccarthy2018defense}.  Even so, data minimization remains one of the principles that ought to be respected regardless of the specific context of personal data processing. Controversies around the principle will continue as the adoption of the draft Data Governance and Data Acts would force discussions as to how to reconcile related legislative incentives to process more (personal) data with data minimization. In this paper, we discuss whether it is at all possible to reconcile data minimization with machine learning. 

Article 5(1)(c) GDPR provides that data shall be ‘adequate, relevant and limited to what is necessary in relation to the purposes for which they are processed’. Article 25(2) GDPR reiterates that controllers only process personal data ‘necessary for each specific purpose of the processing’. First, the data processed must be ‘relevant’, meaning that only pertinent data ought to be processed.  This is designed to safeguard against the accumulation of data for the sake of gathering more data for undisclosed ends and stands in tension with the contemporary operation of ML systems, which often re-purpose data. Second, data can only be processed where it is adequate. Although this requirement is closely intertwined with relevance, adequacy is different in that it may sometimes require that more, not less, data is processed, such as where existing data is inadequate to draw inferences about demographic groups underrepresented in a dataset.  Third, only necessary data ought to be processed, meaning that data controllers need to identify the minimum amount of data necessary to fulfill the purpose~\cite{finckreviving}.  Beyond, there remain unresolved questions regarding the interpretation of data minimization, such as whether data minimization also requires the pseudonymisation of personal data and whether it implies that preference should be given to ordinary personal data over sensitive data~\cite{finckreviving}. Either way, it is worth pointing out that compliance with data minimization can also enhance the quality of ML as there is less need to clean the data and less risk of inaccuracy~\cite{norwegianAuthorityArtificial}.

Data minimization in data-driven systems has been hindered by the lack of concrete computational formalizations, as the principle has received less academic attention in the computing community compared to fairness or transparency. While legal requirements leave room for interpretation, regulatory bodies then issue more specific guidelines to help translate these requirements into practice (guidelines for the implementation of data minimization in Machine Learning and Artificial Intelligence have been issued by the data protection authorities in the UK~\cite{icoDataMinimizationTechniques,icoDataMinimization,icoDataProtectionConcepts} or Norway~\cite{norwegianAuthorityArtificial}). Often, these guidelines mention potential techniques or implementation directions but are still not concrete enough to offer specific mathematical definitions or algorithms (which may lead to vastly varying implementations in practice~\cite{senarath2018understanding}). The paper addresses this gap, exploring the feasibility of an interpretation based on algorithmic performance curves. 

%% file: 09_related_work.tex
\section{Related Work}
\label{Relatedwork}

 Ideas related to the legal concept of data minimization exist across fields in machine learning. We discuss past work on data minimization and performance curves below, and expand on intersections with the literature on sample complexity and active feature acquisition in \vagueref{Appendix \ref{appendix:related_work}}.
 






%

%

\subsection{Perspectives on Data Minimization.} 

Our work follows a number of recent efforts to formalize the principle of data minimization. \citet{biega2020dm} propose an interpretation of data minimization that ties data collection purpose to system performance and discuss the feasability of data minimization in recommendation systems. Our work deepens this interpretation, noting that past work on privacy by design highlights data collection as a key area to implement data minimization. More specifically, we propose a learning-based framework to enforce a data collection stopping criterion, in addition to novel definitions of stopping criteria related to the returns on additional data, rather than absolute model performance. 

Existing guidelines distinguish between \emph{breadth-based} data minimization and \emph{depth-based data minimization} \cite{artintpriv}. In the former, one aims to minimize the number of features, while the latter concerns minimizing the overall amount of data collected for one data modality. \citet{rastegarpanah2021auditing} study breadth-based data minimization and propose an audit method that uses feature imputation to identify whether the features used for a given model are necessary to preserve predictive performance to a pre-specified degree. \citet{goldsteen2021data} also address breadth-based data minimization by identifying how to best generalize features during inference, using ideas from knowledge distillation and data anonymization. In contrast, FIDO is a depth-based data minimization framework meant to guide data collection during model training.

While it has been shown empirically that data can be minimized through various domain- and algorithm-specific heuristics~\cite{biega2020dm}, a question remains of how to automatically learn when to stop data collection for various personalization systems and feature acquisition strategies; the remainder of this paper focuses on this problem. For a thorough treatment of the harms data minimization protects against, we refer the interested reader to \citet{finckreviving}.

\subsection{Performance Curves.} 

Our approach is closely related to empirical research on performance curve estimation, which examines the relationship between dataset size and model performance. The literature considers many metrics, including sensitivity \cite{HAJIANTILAKI2014193}, error rates \cite{gurses2015engineering}, accuracy \cite{DBLP:journals/corr/ChoLSCD15, kolachina2012prediction}, and confidence \cite{Dobbin108, kalayeh1983predicting}. While these works typically assume a power law relationship,  alternatives have been considered and shown to be comparable in accuracy \cite{domhan2015speeding, kolachina2012prediction}.  

\citet{tae2020slice} propose a data collection framework most closely related to ours. They use performance curves to identify classes which require more data to achieve equitable error rates. \citet{tae2020slice} assume a power law relationship throughout the data collection process. In contrast, we model regions of the performance curve separately, and most importantly, use the performance curve to \emph{identify a data collection stopping point}.

Literature on learning curves---the relationship between training epochs and model performance---is related to our own, but aims to learn curves that are generalizable across model configurations or hyperparameter sets. Thus, the methods assume access to hundreds of architectures \cite{baker2017accelerating} or multiple datasets \cite{wistuba2020learning} and depart from our setting. We include a nonparametric baseline inspired by this work in \vagueref{Appendix \ref{sec:nonparametric-regression}} and find that the resulting performance curves are insufficient for achieving data minimization criteria.

%% file: 02_2_problem_statement.tex
\section{Problem Formalization}

\begin{figure*}[t!]
    \centering
    \includegraphics[width=\textwidth]{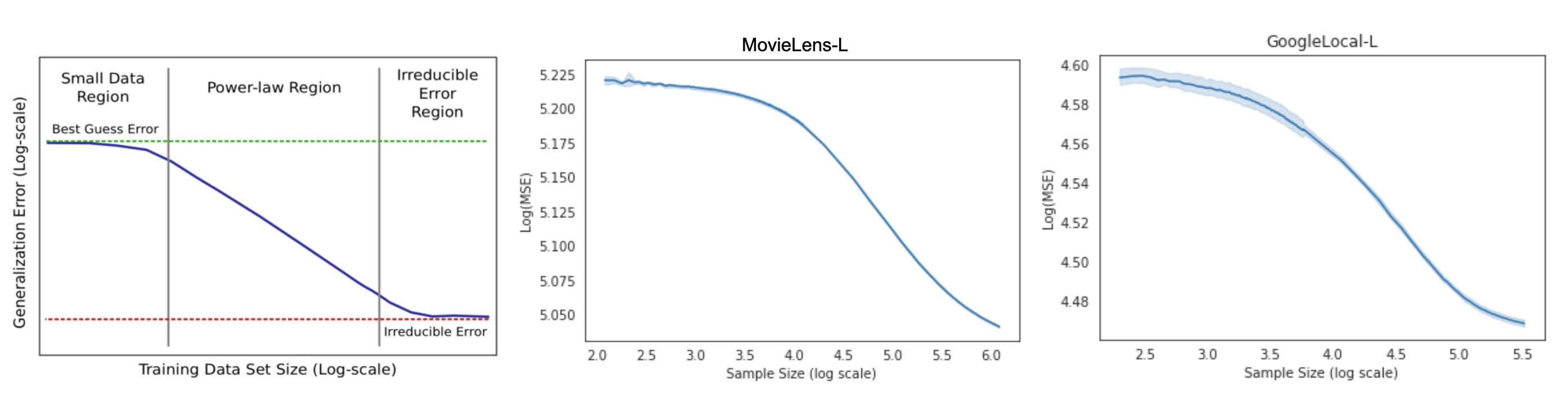}
    \caption{Model performance over the course of data collection. On the left, a figure from \citet{hestness1712deep} plots the phases of data collection. On the right, we plot model performance over data collection from GoogleLocal and MovieLens-20M. Note the change in slope over the course of data collection; we see for both GoogleLocal and MovieLens-20M, the return on additional data decreases, in line with current understanding of the diminishing returns region. Preprocessing details for each dataset can be found in \vagueref{Appendix \ref{sec:data-pipeline}}.}
    \label{perf_curves}
    \Description{The figure is a three panel figure, where the first panel represents the theoretical relationship between dataset size and performance, and the other panels are validation of this relationship on two popular recommendation datasets. On the left, the curve is flat at the beginning, steep in the middle, and flat again at the end. These phases of the performance curve correspond to different stages of data collection, labeled as the ``small data region", the ``power-law region" and the ``irreducible error region. On the right, we see each of these phases occur over the course of data collection on MovieLens (middle) and GoogleLocal (right).}
\end{figure*}

The key proposal in this paper is to maximally limit data collection given a target performance using \emph{performance curves}. We plot the true performance curves for the GoogleLocal-L dataset \cite{pasricha2018translation} and MovieLens-20M dataset \cite{harper2015movielens} and contrast these to the three phases of data collection identified by \citet{hestness1712deep} (Fig. \ref{perf_curves} left, courtesy of \citet{hestness1712deep}). The phases are:

\begin{enumerate}
    \item \emph{The small data region}, where the collected data is insufficiently representative and model performance is poor.
    \item \emph{The power-law region}, where there is a direct trade-off between the amount of data collected and performance.
    \item \emph{The irreducible error/diminishing returns region}, where the collection of more data does not lead to model improvement.
\end{enumerate}

    We can make two observations from Figure \ref{perf_curves}. The first is that the phases of data collection identified by \citet{hestness1712deep} in machine translation exist for recommendation datasets. This is true for both GoogleLocal and MovieLens. The second observation is that a mere 10\% of the dataset lands data collection outside of the small data region. A majority of data collection occurs between the power law region and the diminishing returns region. This has important implications for modeling the performance curve: modeling the entire region using one power law curve produces underestimates of the predicted generalization error given additional data. In later sections, we show that modeling each phase separately mitigates this effect.

From a practical perspective, the data collection phases could be used to decide when collecting more data is necessary for reliable model performance (the small data region), when a user should opt to trade more data for better performance (the power-law region), and when data collection should stop (the irreducible error, or diminishing returns region).

The distinction of these phases is also pertinent from a legal perspective. In particular, the application of data minimization's necessity criterion would indicate that continued collection of personal data in the third phase would be hard to justify as it is not ``necessary" to improve the model and meet its underlying purpose. We formalize these implications into a formal stopping criterion based on an empirical derivative of the the learned performance curve.


\subsection{Formal Interpretation.}
We follow a recent interpretation that ties data collection to \emph{model performance metrics}~\cite{biega2020dm}. However, operationalizing this interpretation remains unclear. We propose a formalization based on the returns in performance from additional data.


\subsubsection{Scenario and Notation.} We assume a scenario where a data processor operates a service (a model $\dmModel$, such as a recommender system) and collects data from a pool of queryable data $\dmPool$ (consisting of user-feature-value triples) generated by a population of users $\dmUsers$.  We define $\dmModel$ as the collection of parameters learned using personal data, including hyperparameters. 

The acquired data is used to train $\dmModel$ and make predictions for each user $\dmUser \in \dmUsers$. We further assume that, when data collection begins, the data processor has access to some initial data: $\dmTraining$ for training the model and $\dmValidation$ set aside to validate model performance predictions. Such initial data would include any data that is historical, purchased, or collected in different markets.

During data collection, the processor applies a feature acquisition policy $\dmPolicy(\dmPool, \dmBudget)$ which queries $n$ feature values from $\dmPool$.  Queries equate to the collection of a specific user-feature-value for inclusion in the training set for model $\dmModel$. We refer to the union of initial and acquired data as $\dmHistoricData$ and let $|\cdot|$ denote the cardinality of a set. Let $\dmPerformance$ represent the true performance curve for $\dmModel$, which maps a domain of training dataset sizes to a range of model performances as measured by a performance metric $\sigma$. User-specific performance curves for $\dmModel$ are termed $\dmUserPerformance$ for a given user $\dmUser$. The objective used to train $\dmModel$ translates to the processing purpose.

The processor can use the resulting predicted performance curve to adhere to a concrete data minimization objective. In the main text, we propose and validate one such objective, detailed below: \emph{returns-based} data minimization. Experiments in \vagueref{Appendix \ref{sec:alternate-dm-objective}} furthermore engage an alternate formalization of data minimization, where the stopping criterion is determined by the \emph{relative model performance} achieved rather than the performance returns, providing further evidence of FIDO's flexibility.

\subsubsection{Minimizing by Returns in Performance.}  We minimize in reference to a threshold on the \textit{return} in model performance of additional data. One could select an appropriate threshold by assessing user preferences or selecting a  sufficiently small threshold such that user experience is not affected. Formally, a data collector would cease data collection once the slope of the performance curve drops below threshold $t$: 
\begin{equation}
    \frac{d\dmPerformance}{dn}(|\mathcal{A}|) \leq t
\end{equation}
Producing an accurate approximation of $\dmPerformance$ is thus central to any performance-based data minimization objective. Our experiments show that existing approaches produce performance curves that are insufficient for the stated objectives. We compensate for these shortcomings by providing an accurate parametric model to approximate $\dmPerformance$.

%% file: 04_method_three_stage.tex
\section{FIDO: \textbf{F}ramework for \textbf{I}nhibiting \textbf{D}ata \textbf{O}vercollection}

The framework accepts three parameters: feature acquisition algorithm $\dmPolicy$, model $\dmModel$, and performance metric $\sigma$.  
There are three steps: (1) $\dmPolicy$ acquires a portion of the available data (\emph{Data Collection}), (2) the performance curve is fit to the new data (\emph{Curve Fitting}), and (3) Steps 1 and 2 repeat until the conditions of Step 3 (\emph{Stopping Criterion Evaluation}) are met.

\subsection{Step 1: Collect Data.} In this step, a feature acquisition algorithm $\dmPolicy$ collects $\dmNumQueries$ observations from the pool of available observations $\dmPool$. Smaller $\dmNumQueries$ translate to more conservative data collection processes and to more accurate estimates of the stopping criterion at the expense of decreased efficiency (smaller $\dmNumQueries$ mean that Steps 2 and 3 are executed more frequently). One might choose to set a larger $\dmNumQueries$ early on during data processing, and decrease it as the data processing continues. 

\subsection{Step 2: Fit the Performance Curve.}
\label{sec:ppl-technique}

The key idea underpinning this step is to model the phases of data collection \textit{separately} via a piecewise power law curve. The piecewise power law curve is the piece-wise combination of three power law curves, which we will refer to as $\dmLowerBound$, $\dmMiddleBound$, and $\dmUpperBound$:
\begin{equation}
f(x) = \begin{cases} 
      f_0(x) = a_0x^{-b_0} & 0 \leq x \leq t_0 \\
      f_1(x) = a_1x^{-b_1} & t_0 < x  \leq t_1\\
      f_2(x) = a_2x^{-b_2} & t_1 < x \\
   \end{cases}
\end{equation} 
where $f(x)$ accepts as input a training set size $x$. We fit the piecewise power law curve to subsamples of $\dmHistoricData$ of different sizes. More specifically, given the query size parameter $\dmNumQueries$, we generate \( |\dmHistoricData| / \dmNumQueries\)  samples such that the size of each consecutive sample increases by $q$. We train the model on each sample and evaluate model performance on $\dmValidation$. The resulting pairs of values (sample size and performance on the validation set) are then used to fit $f(x)$.  We fit the parameters for $\dmLowerBound$, $\dmMiddleBound$, and $\dmUpperBound$ using weighted non-linear least squares; details can be found in \vagueref{Appendix \ref{sec:parameter-fitting}}.


Finally, we optimize thresholds $t_0$ and $t_1$ using coordinate descent, such that the \emph{slopes} of each consecutive pair of power laws maximally differ. This translates to an iterative approach that first estimates $t_0$ while $t_1$ is left fixed, and then estimates $t_1$ while $t_0$ is fixed, until convergence. More formally, we alternate optimization between the following objectives: 

\begin{eqnarray}
    \max_{t_0} \quad|b_0 - b_1| \quad &s.t.& \quad 0 < t_0 < t_1\\
    \max_{t_1} \quad |b_1 - b_2| \quad &s.t.& \quad t_0 < t_1 < |\dmHistoricData|
\end{eqnarray}

While $t_0$ and $t_1$ do not appear directly in the optimization objective, they impact the estimates of $b_0$, $b_1$, and $b_2$ by delimiting which sample size and model performance pairs are used to estimate each decay parameter (as described in Equation (2)). Assuming the underlying function is a piecewise power law curve, Equations (3) and (4) are convex optimizations and converge to the correct thresholds.

This follows our intuition regarding the each region: namely, that the phases are distinguished by the differing \emph{return} in additional data. Note that in this paper we assume that model performance increases as we collect more data. This is a common assumption in the performance curve literature \cite{kolachina2012prediction, hestness1712deep}, but there are cases in which additional data may hurt model performance. We discuss one such case in Section \ref{sec:user-specific}. In these settings, a different family of parametric curves should be used and FIDO remains flexible to alternate performance curve estimation procedures. 

In theory, one could compute a stopping criterion directly from the subsamples of $\mathcal{A}$, without fitting the performance curve. This has two undesirable consequences. The first is the instability of the resulting stopping criteria, due to the noise inherent to performance measurements from individual subsamples, as we will see in later experiments. Second, learning a performance curve allows the data minimizer to reason about performance given additional data, \emph{without collecting that data}. This affords the framework flexibility to a range of data minimization objectives, including those that cease data collection based on absolute model performance rather than performance increase rate (relevant experiments are in the supplement).








\subsection{Step 3: Evaluate Stopping Criterion.} In this step, the resulting performance curve is used to accomplish a specific data minimization objective. Note that these are not the only reasonable data minimization objectives and the framework can adapt to different formulations. 

Minimizing by returns requires the data collector to specify a threshold for return after which data collection should stop, $\dmSlopeThreshold \in \mathbb{R}$. We can use the performance curves to estimate this quantity by taking the derivative at a given sample size: 
\begin{equation}
\dmSlopeEstimate = \begin{cases} 
      -b_0 a_0x^{-b_0-1} & 0\leq x \leq t_0 \\
      -b_1 a_1x^{-b_1-1} & t_0 < x \leq t_1 \\
      -b_2 a_1x^{-b_2-1} & t_1 < x  \\
   \end{cases}
\end{equation}
Once $\hat{s}$ falls below $\dmSlopeThreshold$, data collection stops. Implementation details are in \vagueref{Appendix \ref{sec:implementation-details}}.




%% file: 05_experiment_setup.tex
\section{Experiments}



\input{05_01_dataset_statistics}

\subsubsection{Datasets.} We perform experiments on two datasets in the recommender system domain: MovieLens-20M \cite{harper2015movielens} and GoogleLocal \cite{he2017translation, pasricha2018translation}. The datasets contain user ratings for movies and businesses, respectively. For each, the task is to predict user ratings for unseen items. We sample each dataset at two sizes to examine how results generalize across user numbers and sparsity levels. Dataset statistics can be found in Table \ref{dataset_info} and preprocessing pipelines can be found in \vagueref{Appendix \ref{sec:data-pipeline}}.

Each dataset is subject to the same initial, validation, and test splits, where each split is 10\% of the total ratings and stratified across users. The remaining 70\% of the data is the queryable rating set $\dmPool$. We produce 5 random splits of each dataset according to these divisions. All results are reported over the 5 splits. We assume random feature acquisition unless otherwise stated.

 \subsubsection{Alternate Curve Models.} Methods relating dataset size to performance commonly assume a power law model \cite{hestness1712deep, tae2020slice}. We benchmark FIDO's piecewise power law technique against alternate approaches in the literature. We include \textit{2P-PL-Initial} to determine the benefit of updating the curve as data is acquired, and a two-parameter power law method \textit{2P-PL} to represent the most common approach to fitting performance curves \cite{tae2020slice, Figueroa_2012}. The remaining baselines represent variations of the power law curve that capture the  notion of diminishing returns. The first (\textit{3P-PL}) models the irreducible error  directly and  the second  (\textit{3P-PL-Exp}) models an additional exponential decay. The parameter fitting approach is the same for our method and described in \vagueref{Appendix \ref{sec:parameter-fitting}}. 

\begin{itemize}
\itemsep 0em
    \item \textit{2P-PL-Initial}: Fits two-parameter power law ($f(x) = ax^b$)  to subsamples of $\dmTraining$.
    \item \textit{2P-PL}: Fits two-parameter power law ($f(x) = ax^b$) to subsamples of $\dmHistoricData$. 
    \item \textit{3P-PL}: Fits three-parameter power law ($f(x) = ax^b+c$) to subsamples of $\dmHistoricData$. 
    \item \textit{3P-PL-Exp}: Fits three-parameter power law with an exponential cutoff ($f(x) = x^a e^{bx} + c$) to subsamples of $\dmHistoricData$.
    \item \textit{Naive}: Estimates slope of the performance curve empirically via last two subsamples of $\mathcal{A}$.
    \item \textit{Oracle}: Estimates slope via a discrete approximation using \emph{all} sample size and performance pairs. Exact implementation is in \vagueref{Appendix \ref{sec:oracle-implementation}}.
\end{itemize}

We include comparisons to two additional baselines---a variant of the proposed curve model with two pieces (for the power law curve stage and diminishing returns stage) rather than three, and a nonparametric regression model---in \vagueref{Appendix \ref{sec:nonparametric-regression}}.

\subsubsection{Hyperparameters.} We assume that $\dmModel$ is a FunkSVD \cite{funk2006netflix} recommendation system. We use the same hyperparameters for the number of latent features $r$ and query size $q$ across all experiments: $r$ is set to be 30, and $q$ is set to be 2\% of the number of queryable entries.

\subsubsection{Metrics.} We measure performance via mean-squared error (MSE), a standard recommendation evaluation metric. Methods are compared based on (1) return given additional data (change in MSE per additional feature-value observation) and (2) the amount of data collected for a given threshold $t$ (reported over 5 dataset splits). We compute statistical significance via a paired $t$-test with a Bonferroni correction.

%% file: 05_01_dataset_statistics.tex
\begin{table}
\centering
\small
\resizebox{7cm}{!}{
    \begin{tabular}{ |c|c|c|c|c| } 
    \hline
     \textbf{Dataset} & \textbf{\# Users} & \textbf{\# Items} & \textbf{Item type} & \textbf{Sparsity} \\
     \hline
      MovieLens-L & 5000 &  17400 & movie & 1.7\% \\
     MovieLens-S & 1000 & 11529 & movie & 2.6\% \\ 
     GoogleLocal-L  & 1500 & 265807 & business & 0.1\% \\ 
     GoogleLocal-S  & 500 & 104766 & business & 0.3\% \\
     \hline
    \end{tabular}}
\caption{Dataset statistics.}
\vspace{-1cm}
\label{dataset_info}
\end{table}


%% file: 06_experiments.tex
\subsection{Evaluation of Data Minimization Objective.}




We compare the amount of data collected by FIDO using a suite of performance curve models, including the piecewise power law technique described in Section \ref{sec:ppl-technique}, to the amount of data collected by an oracle with access to \emph{all} sample size and performance pairs. For a variety of thresholds, FIDO paired with the piecewise power law technique halts data collection significantly closer to a criterion with access to all sample size and performance estimates (Table  \ref{diminishing_returns_table}, $p$ < 1e-5). Due to the noise of the performance measurements, \emph{Naive} produces stopping criteria that are both noisier and further from the true stopping point. Later experiments show that \emph{Naive} fails to generalize to different feature acquisition algorithms and cannot accommodate alternate performance-based data minimization objectives.

\begin{figure*}[t!]
    \centering
    \includegraphics[width=0.93\textwidth]{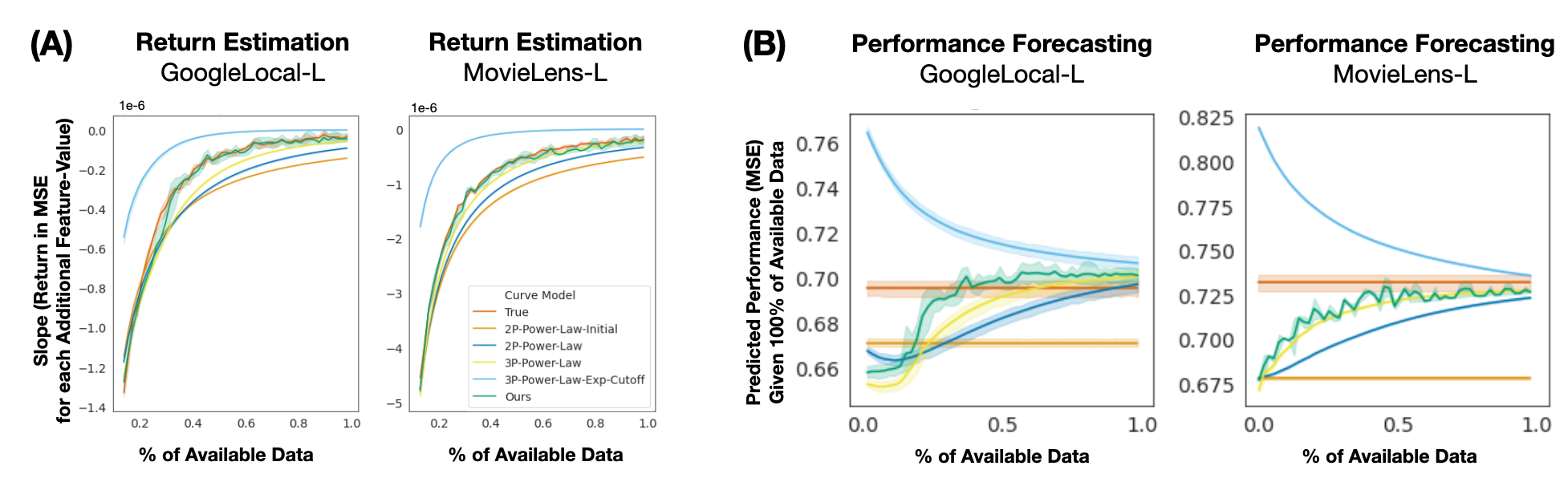}
    \caption{\textbf{Evaluation of performance curves over the course of data collection.} Our method outperforms baselines in estimating \emph{return}  on additional data, where return is defined as the reduction in model error given an additional feature-value observation \textbf{(A)}. Supporting plots for GoogleLocal-S and MovieLens-S are in \vagueref{Appendix \ref{sec:other-datasets}}. For each dataset, we plot predicted performance, $\dmPerformance{(|\dmUnion{\dmTraining}{ \dmPool}|)}$, over the course of data collection using different curve-fitting methods \textbf{(B)}. Our method (dark green) matches the true performance (red) most closely at all stages.}
    \label{diminishing_returns_fig}
    \Description{The figure has four panels, where the left two panels plot the relationship between dataset size and estimated return on additional data and the two panels on the right plot estimated performance given additional data. The baselines are plotted in different colors on each graph. Over both estimation tasks, FIDO combined with a piecewise power law curve achieves the highest accuracy.}
\end{figure*}

The slope estimates drawn from the performance curve in FIDO are more accurate than those of the curve-fitting baselines across the stages of data collection (Figure \ref{diminishing_returns_fig} (A)). \emph{3P-PL-Exp} consistently underestimates the return on additional data over the course of data collection, and subsequently halts data collection too early. In contrast, \emph{2P-PL-Initial},  \emph{2P-PL}, and \emph{3P-PL} overestimate the return on additional data and frequently halt data collection later than the empirically derived stopping point. These results suggest  that the use of \emph{3P-PL-Exp} would produce a conservative data minimization approach in that such a  data minimization method would be unlikely to over-collect data. The reverse is true for the remaining baselines -- such a data minimization approach would likely collect more data than is required.

This is a direct result of each curve model's ability to model the last stage of data collection accurately. Examine the accuracy of each curve model's estimate of performance given the entire dataset (Figure \ref{diminishing_returns_fig} (B)). Each method converges to the true value for model performance (red) over the course of data collection. The power law baselines (\emph{2P-PL-Initial}, \emph{2P-PL}, \emph{3P-PL}) underestimate error  given $\dmPool$. This confirms results from prior work in machine translation \cite{kolachina2012prediction}, and is  a consequence of extrapolation from the power law region into the diminishing returns region. \emph{3P-PL-Exp} instead overestimates test error because the $e^{-bx}$ term produces a curve too flat to describe the true relationship; illustrative plots for the performance curve fits are included in \vagueref{Appendix \ref{sec:pc-fit-illustrations}}.



The halting points for GoogleLocal-S and MovieLens-S exhibit more noise than their larger counterparts. This suggests that producing reliable estimates of the return on additional data is more challenging for smaller datasets.

\input{06_01_table}

\subsection{Robustness to Query Size.} In the previous experiments we used a query size of 2\%. Here, we investigate the robustness of FIDO's predictions to different query size values $\dmNumQueries \in [0.005, 0.01, 0.02, 0.03, 0.04, 0.05, 0.06, 0.07]$. 
Large $q$ simulate a setting in which large batches of data are acquired at one time, while smaller $q$ simulate the continuous arrival of new user data.

Table \ref{query_size_table} reports that FIDO provides the closest estimation of the true model performance across different query sizes. While FIDO produces the most  faithful estimates of the true stopping point across query sizes \emph{it is more sensitive to small query sizes} compared to baselines using a single power law curve. Lower query sizes require models to be retrained more frequently to produce data to fit the performance curve. Thus, for models with computationally intensive  training procedures, it is optimal to choose the largest query size that maintains accuracy. On the other hand, smaller queries will lead to more accurate stopping decisions and less data overcollection. It is up to a practitioner to select the query size based on their domain knowledge, and our results suggest that the set of query sizes producing accurate estimates is quite large.

\subsection{Robustness to Feature Acquisition Algorithms.} Thus far, we have considered data collection where observations are queried randomly from $\mathcal{Q}$. AFA methods improve upon this approach by instead querying feature values based on their uncertainty \cite{huang2018active,freund1997selective,chakraborty2013} or contribution to a downstream task \cite{melville2005expected, Vu_intelligentinformation}. Successful AFA methods collect less data than random feature acquisition and deliver equivalent performance. Recent work has shown that this success is often dependent on initialization conditions \cite{munjal2020towards}. 

We consider two popular AFA methods: \textit{Stability} and \textit{Query-by-Committee} (\emph{QBC}). \textit{QBC} \cite{chakraborty2013} employs three matrix imputation approaches ($k$-NN, EM, and SVD) to predict missing feature values. \emph{Stability} takes a similar approach and predicts missing feature-values using SVD given different ranks. Each feature-value's uncertainty corresponds to the variance in predicted values. Both algorithms request the highest variance feature-values. For \textit{Stability}, we follow the approach of \cite{huang2018active} and set the ranks to be $[1, 2, 3, 4, 5]$. 

\subsubsection{FIDO is robust to different AFA algorithms.}

Across both AFA algorithms and multiple thresholds, FIDO halts data collection closest to the true stopping point ($p$ <1e-3). Table \ref{afa_table} reports these results for GoogleLocal-L. Expectedly, we see that for the same stopping criterion (e.g., a threshold of -5.0e-7), AFA algorithms collect less data than random feature acquisition algorithms. Existing curve-fitting approaches are more competitive in the AFA setting, which can be attributed to an expanded power law region (illustrative performance curves can be found in \vagueref{Appendix \ref{sec:afa-perf-curves})}. 


In general, trends for each baseline hold in this setting: the curve models which use a single power law (\emph{2P-PL-Initial}, \emph{2P-PL}, \emph{3P-PL}) stop collecting data too late on average, while \emph{3P-PL-Exp} stops collecting data too early. \emph{Naive} is naturally affected by the noisier performance curves and performs significantly worse than the other methods. 

\subsubsection{AFA algorithm performance depends on initial system data} In Figure \ref{user_considerations}(A) we examine the dependence of AFA algorithm performance on (1) the type of initialized data and (2) the feature acquisition algorithm employed. We consider two additional types of initialization; user-subset (initialized randomly across a subset of users) and item-subset (initialized randomly across subset of items). In each of these cases, the test set is formed from a random sample that includes ratings from all users. In agreement with \citet{munjal2020towards}, we observe that the performance depends on data initialization conditions. When the initialization data is a random sample across users and items, AFA algorithms perform similarly. However, when the initialization data contains only a subset of users, or only a subset of items, AFA begins to decrease in performance. This is consequential in cases where (i) the population of data subjects is evolving (initialization data does not contain users who join at a later time), and (ii) the data processor is not initially allowed to collect certain feature values because of external constraints (e.g., feature sensitivity).

\input{06_02_step_size_table}

\input{06_03_afa_table}



%% file: 06_01_table.tex
\begin{table*}
\centering
\tiny
\resizebox{1\textwidth}{!}{%

\begin{tabular}{l|c|cccccc|c}
\toprule
      Dataset & Threshold & 2P-PL-Initial &         2P-PL &         3P-PL &     3P-PL-Exp &         Naive &          FIDO &          Oracle \\
\midrule
GoogleLocal-L &  -5.0e-07 & $0.32 \pm 0.00$ & $0.32 \pm 0.01$ & $0.32 \pm 0.01$ & $0.16 \pm 0.00$ & $\mathbf{0.27 \pm 0.02}$ & $\mathbf{0.29 \pm 0.02}$ & $0.27 \pm 0.01$ \\
GoogleLocal-L &  -2.0e-07 & $0.73 \pm 0.01$ & $0.61 \pm 0.01$ & $0.53 \pm 0.01$ & $0.25 \pm 0.01$ & $0.36 \pm 0.04$ & $\mathbf{0.42 \pm 0.03}$ & $0.41 \pm 0.02$ \\
GoogleLocal-L &  -5.0e-08 & $1.00 \pm 0.00$ & $1.00 \pm 0.00$ & $1.00 \pm 0.00$ & $0.40 \pm 0.02$ & $0.52 \pm 0.08$ & $\mathbf{0.68 \pm 0.05}$ & $0.68 \pm 0.05$ \\
\midrule
GoogleLocal-S &  -5.0e-07 & $0.88 \pm 0.01$ & $0.71 \pm 0.01$ & $0.60 \pm 0.01$ & $0.28 \pm 0.01$ & $0.37 \pm 0.07$ & $\mathbf{0.42 \pm 0.06}$ & $0.47 \pm 0.03$ \\
GoogleLocal-S &  -2.0e-07 & $1.00 \pm 0.00$ & $1.00 \pm 0.00$ & $0.91 \pm 0.03$ & $0.39 \pm 0.01$ & $0.44 \pm 0.08$ & $\mathbf{0.51 \pm 0.14}$ & $0.58 \pm 0.06$ \\
GoogleLocal-S &  -5.0e-08 & $1.00 \pm 0.00$ & $1.00 \pm 0.00$ & $1.00 \pm 0.00$ & $\mathbf{0.55 \pm 0.01}$ & $0.46 \pm 0.09$ & $\mathbf{0.59 \pm 0.08}$ & $0.64 \pm 0.10$ \\
\midrule
  MovieLens-L &  -5.0e-07 & $0.13 \pm 0.00$ & $0.13 \pm 0.00$ & $0.13 \pm 0.00$ & $0.13 \pm 0.00$ & $\mathbf{0.14 \pm 0.01}$ & $\mathbf{0.13 \pm 0.00}$ & $0.13 \pm 0.00$ \\
  MovieLens-L &  -2.0e-07 & $0.29 \pm 0.00$ & $0.27 \pm 0.00$ & $0.26 \pm 0.00$ & $0.13 \pm 0.00$ & $\mathbf{0.25 \pm 0.01}$ & $\mathbf{0.25 \pm 0.02}$ & $0.23 \pm 0.01$ \\
  MovieLens-L &  -5.0e-08 & $1.00 \pm 0.00$ & $0.76 \pm 0.01$ & $0.62 \pm 0.01$ & $0.24 \pm 0.01$ & $0.43 \pm 0.08$ & $\mathbf{0.53 \pm 0.05}$ & $0.53 \pm 0.02$ \\
  \midrule
  MovieLens-S &  -5.0e-07 & $0.50 \pm 0.02$ & $0.53 \pm 0.01$ & $0.51 \pm 0.01$ & $0.23 \pm 0.01$ & $\mathbf{0.37 \pm 0.04}$ & $\mathbf{0.46 \pm 0.05}$ & $0.45 \pm 0.03$ \\
  MovieLens-S &  -2.0e-07 & $1.00 \pm 0.00$ & $1.00 \pm 0.00$ & $0.97 \pm 0.02$ & $0.36 \pm 0.01$ & $0.51 \pm 0.08$ & $\mathbf{0.68 \pm 0.14}$ & $0.79 \pm 0.08$ \\
  MovieLens-S &  -5.0e-08 & $1.00 \pm 0.00$ & $1.00 \pm 0.00$ & $1.00 \pm 0.00$ & $0.64 \pm 0.02$ & $0.66 \pm 0.22$ & $\mathbf{0.99 \pm 0.03}$ & $1.00 \pm 0.00$ \\
\bottomrule
\end{tabular}

}
\caption{\textbf{Performance over diminishing returns criterion.} Each value is the fraction of data collected using a given method while adhering to the diminishing returns stopping criterion, averaged over 5 runs. Our method halts collection closest to the true stopping point across thresholds (p-value 3e-6). Bolded entries are closest to the true stopping point, within a standard deviation.}
\label{diminishing_returns_table}
\end{table*}

%% file: 06_02_step_size_table.tex
\begin{table*}
\tiny
\centering
\resizebox{1\textwidth}{!}{%

\begin{tabular}{c|c|cccccc|c}
\toprule
 Query Size & Threshold & 2P-PL-Initial &         2P-PL &         3P-PL &     3P-PL-Exp &         Naive &          FIDO &        Oracle \\
\midrule
      0.005 &  -2.0e-07 & $0.70 \pm 0.01$ & $0.62 \pm 0.01$ & $0.55 \pm 0.01$ & $0.21 \pm 0.00$ & $0.24 \pm 0.04$ & $0.30 \pm 0.04$ & $0.35 \pm 0.01$ \\
      0.010 &  -2.0e-07 & $0.71 \pm 0.01$ & $0.62 \pm 0.01$ & $0.55 \pm 0.01$ & $0.22 \pm 0.00$ & $0.32 \pm 0.05$ & $0.36 \pm 0.02$ & $0.39 \pm 0.03$ \\
      0.020 &  -2.0e-07 & $0.73 \pm 0.01$ & $0.62 \pm 0.01$ & $0.53 \pm 0.01$ & $0.25 \pm 0.00$ & $0.39 \pm 0.03$ & $0.41 \pm 0.03$ & $0.42 \pm 0.02$ \\
      0.030 &  -2.0e-07 & $0.75 \pm 0.02$ & $0.63 \pm 0.02$ & $0.53 \pm 0.02$ & $0.27 \pm 0.01$ & $0.43 \pm 0.02$ & $0.46 \pm 0.02$ & $0.43 \pm 0.01$ \\
      0.040 &  -2.0e-07 & $0.79 \pm 0.03$ & $0.64 \pm 0.02$ & $0.53 \pm 0.02$ & $0.29 \pm 0.00$ & $0.44 \pm 0.02$ & $0.46 \pm 0.02$ & $0.44 \pm 0.02$ \\
      0.050 &  -2.0e-07 & $0.81 \pm 0.04$ & $0.65 \pm 0.02$ & $0.54 \pm 0.02$ & $0.32 \pm 0.00$ & $0.45 \pm 0.05$ & $0.49 \pm 0.02$ & $0.46 \pm 0.02$ \\
      0.060 &  -2.0e-07 & $0.78 \pm 0.02$ & $0.63 \pm 0.00$ & $0.52 \pm 0.02$ & $0.32 \pm 0.00$ & $0.45 \pm 0.03$ & $0.47 \pm 0.00$ & $0.45 \pm 0.03$ \\
      0.070 &  -2.0e-07 & $0.84 \pm 0.03$ & $0.64 \pm 0.03$ & $0.50 \pm 0.00$ & $0.35 \pm 0.03$ & $0.49 \pm 0.03$ & $0.50 \pm 0.00$ & $0.47 \pm 0.03$ \\
\bottomrule
\end{tabular}

}
\caption{\textbf{Robustness to Query Size.} 
Applied to GoogleLocal-L, each method's robustness to different query sizes sheds light on the tradeoffs involved in query size selection. The proposed method and \textit{Naive} are competitive in their accuracy in estimating the true stopping point, while \textit{2P-PL} and \textit{3P-PL} produce the most consistent stopping criteria  over threshold sizes.}
\label{query_size_table}
\end{table*}

%% file: 06_03_afa_table.tex

\begin{table*}
\tiny
\centering
\resizebox{1\textwidth}{!}{%
\begin{tabular}{l|c|cccccc|c}
\toprule
  AFA Alg & Threshold & 2P-PL-Initial &         2P-PL &         3P-PL &     3P-PL-Exp &         Naive &          FIDO &          Oracle \\
\midrule
Stability &  -5.0e-07 & $0.30 \pm 0.00$ & $0.31 \pm 0.03$ & $0.31 \pm 0.06$ & $0.16 \pm 0.00$ & $0.24 \pm 0.03$ & $0.26 \pm 0.06$ & $0.22 \pm 0.06$ \\
Stability &  -2.0e-07 & $0.71 \pm 0.02$ & $0.64 \pm 0.06$ & $0.57 \pm 0.11$ & $0.24 \pm 0.03$ & $0.27 \pm 0.09$ & $0.47 \pm 0.11$ & $0.42 \pm 0.09$ \\
Stability &  -1.0e-07 & $1.00 \pm 0.00$ & $1.00 \pm 0.00$ & $0.96 \pm 0.06$ & $0.29 \pm 0.06$ & $0.29 \pm 0.14$ & $0.70 \pm 0.21$ & $0.75 \pm 0.15$ \\
      QBC &  -5.0e-07 & $0.32 \pm 0.02$ & $0.33 \pm 0.00$ & $0.33 \pm 0.00$ & $0.19 \pm 0.02$ & $0.23 \pm 0.00$ & $0.24 \pm 0.02$ & $0.21 \pm 0.02$ \\
      QBC &  -2.0e-07 & $0.73 \pm 0.02$ & $0.67 \pm 0.02$ & $0.58 \pm 0.02$ & $0.24 \pm 0.02$ & $0.23 \pm 0.00$ & $0.38 \pm 0.03$ & $0.36 \pm 0.03$ \\
      QBC &  -1.0e-07 & $1.00 \pm 0.00$ & $1.00 \pm 0.00$ & $1.00 \pm 0.00$ & $0.31 \pm 0.02$ & $0.23 \pm 0.00$ & $0.88 \pm 0.27$ & $0.96 \pm 0.00$ \\
\bottomrule
\end{tabular}
}
\caption{\textbf{Robustness to AFA Algorithm.} The proposed method adheres most closely to the true stopping point across AFA algorithms applied to data collection from GoogleLocal-L.}
\label{afa_table}
\end{table*}

%% file: 08_pitfalls.tex
\label{sec:user-specific}
\section{User-specific Impact Analyses}
Previous sections discuss minimized data collection in terms of diminishing return across all users. In this section, we examine FIDO's effect on per-user metrics. We discuss how user performance-based data minimization departs from traditional assumptions for performance curves and recommend areas for further research.

\subsection{Evaluation of User-Specific Data Minimization.}
\begin{figure*}[t!]
    \centering
    \includegraphics[width=\textwidth]{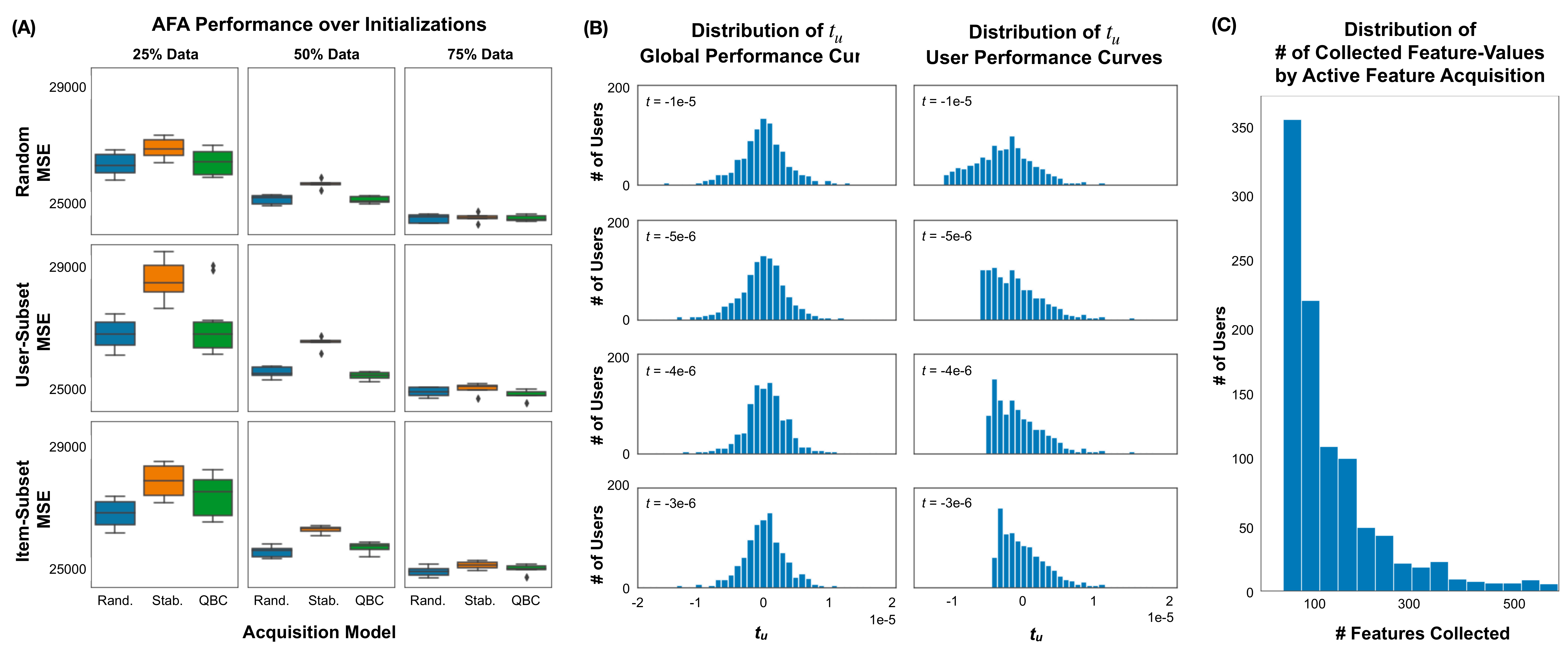}
\caption{\textbf{(A)} We show that the performance achieved during data collection depends on both the AFA algorithm employed and the initialization conditions. Error bars are reported over 5 random initializations. \textbf{(B)} User-specific performance metrics when minimized data collection learns one curve for all users (left) and a curve for each user (right). Unsurprisingly, we see that the mode of user-specific performance increases as $t$ increases. This is also the case when learning user-specific performance curves, but there is a large spread of true returns on additional data. This  suggests future areas of work for learning accurate user-specific performance curves.  \textbf{(C)} We also show that a small portion of users bear the majority of the data collection burden in a histogram of the quantity of features acquired per user by \textit{Stability} from MovieLens-S halfway through data collection.}
\label{user_considerations}
\vspace{.5cm}
\end{figure*}

We analyze the performance of the framework on user-specific performance metrics in two cases. In the first, we replicate the setting discussed in previous sections and learn a performance curve for the entire dataset. In the second, FIDO learns a performance curve \emph{per user} and applies a stopping criterion based on goal $\dmReturn$ to each curve. We use the same procedure to estimate $\dmReturn_\dmUser$ as $\dmReturn$ earlier, except we use a user-specific performance curve rather than a global performance curve. The user-specific performance curve is learned by replacing a global performance metric (MSE over all users) with a user-specific performance metric (MSE over a set of items specific to one user). Estimating $\dmReturn_\dmUser$ from $\dmUserPerformance$ produces a $\dmReturn_\dmUser$ for each user.
    
 In Figure \ref{user_considerations}(B), we plot the distribution of user returns in performance given a global threshold $t$ (left). As $t$ increases, the distribution mode shifts up and the variance in user fractions of performance decreases. Note that the x-axis for each histogram extends beyond $1$. This is because more data does not necessarily translate to increased \emph{per-user} performance. Two factors are responsible: 1) The small validation set size for each user produces noisy performance estimates and 2) the collection of additional data can still hurt user performance if this data is not representative. The assumption of monotonically increasing performance over data collection does not hold, and accordingly, the framework does not perform as well. A key takeaway from this experiment is that the return in performance may not be an appropriate metric for data minimization on a per user level. 

\subsection{Lessons on User-Specific Data Minimization.}

Methods designed to address user-specific data minimization should consider:

\subsubsection{Tensions between AFA and fairness}  While AFA is a natural choice for limited data collection, we find that it introduces disparate data collection burden across users by collecting a different number of features from different users. Figure \ref{user_considerations}(C), plots a histogram of the quantity of collected data over users for AFA algorithm \textit{Stability}, for dataset MovieLens-S (similar trends exist for other datasets). AFA algorithms "exploit" a small number of users by collecting a large number of feature-values from them. Yet, our experiments also show that increased data collection significantly correlates with better performance for individual users. It is likely that certain users would choose to bear the burden of excess data collection in exchange for better performance. Thus, data minimization given an AFA approach raises questions of both user fairness and user agency.

\subsubsection{More complex curve models.} It is worth exploring a family of parametric curves that describe the phenomenon of a user whose model performance degrades with the collection of additional data. Approximately 20\% of users in each of the four datasets exhibit this property, suggesting that monotonic curve models are not the right choice for modeling user-specific performance curves. Moreover, user-specific performance curves are noisier and as a result, may benefit from drawing multiple samples at each subsample size.


%% file: 10_putting_it_all_together.tex
\section{Putting It All Together}
Thus far, we have put forth a framework to guide data collection by learning the an algorithm's performance curve as data arrives. Subsequent experiments validated the resulting performance curve on a returns-based data minimization objective. These experiments deliver a number of takeaways for the data minimization practitioner. Here, we summarize the implications of our results on specific design choices:

\begin{enumerate}
\item \textbf{Choosing a Feature Acquisition Algorithm.} Based on our experiments, we recommend random feature acquisition to guide data minimization. We base this recommendation on three reasons: 1) random feature acquisition produces a smoother performance curve, and thus, more accurate performance estimates, 2) our work confirms recent findings that demonstrate how the success of intelligent data collection depends upon initialization conditions \cite{munjal2020towards}, and 3) AFA algorithms can place excess data collection burden on specific users, as our user-specific impact analyses show.

\item \textbf{Choosing a Performance Curve Model.} The piecewise power law curve model is best fit to describe the relationship between dataset size and performance in the datasets we consider. We see that this choice may not be consequential when the diminishing returns region is small, as we see with MovieLens-S.

\item \textbf{Creating a Representative Validation Set.} This framework hinges on the creation of a representative validation set. As is the case with many systems that operate on continuously collected data, a representative validation set may need to be updated to account for data drift over time.

\item \textbf{Identifying a Relevant Objective.} While the framework we propose is flexible to many objectives, it is worth considering when one objective may be more desirable than another. We provide a case study of an alternate objective, based on relative performance in the supplement. One can consider its \emph{ease of definition}: how well does the selected objective reflect user preferences? Gauging user preferences in terms of relative performance may be easier to survey for than diminishing returns. One could also consider the objective's \emph{specificity}. Our experiments show that minimizing by diminishing returns is more accurate earlier on during data collection compared to minimization by relative performance. Dataset size plays a role too, as our experiments show that both the prediction of returns and relative performance are noisier in smaller datasets.

\item \textbf{Performing User-Specific Impact Analyses.} Data minimization may disproportionately affect marginalized populations. Recent work \cite{tae2020slice, chen2018my} has shown that in some instances, it is necessary to collect \emph{more} data to ensure the equitable performance. As a result, it is imperative to perform per-user analyses when studying and proposing methods for data minimization.  
\end{enumerate} 

%% file: 11_limitations.tex
\section{Limitations}

FIDO is not without limitations. First, the framework relies on access to a validation set large enough to approximate performance on the test set. Experiments on MovieLens-S and GoogleLocal-S show that smaller validation set sizes translate to higher variance in data minimization performance. One could mitigate this variation in performance by intelligently constructing the validation set, rather than randomly. Such an approach could be useful in settings where concept drift is common, and the validation set must be updated to remain representative of the test set.

Moreover, this work considers data acquisition where feature-values are roughly homogeneous in terms of sensitivity, since each is an item rating. Data minimization concerns not only whether data is collected, but also what \textit{type} of data is collected. This includes the identifiability or pseudonymity of the collected data. Intersections between FIDO and recent work on breadth-based data minimization \cite{rastegarpanah2021auditing, goldsteen2021data} can address feature-specific data minimization concerns.

Finally, FIDO is only one way to implement data minimization and represents one facet of a comprehensive approach to data minimization. In practice, a data processor would deploy multiple techniques that work in concert to minimize data across stages of a system's life cycle, including data collection, storage, and inference. 


%% file: 12_conclusion_old.tex
\section{Discussion}
\label{conclusion}

This work aims to bridge the gap between the legal principle of data minimization and its practical realization. Data-driven systems often operate under the assumption that more data is unequivocally better, for all parties involved. This might not the case when the identifiability, sensitivity, liability, and storage costs of collected data are acknowledged and appropriately balanced. We intend our work to be a step towards respectful data collection. 

Towards this end, we propose FIDO, a Framework for Inhibiting Data Overcollection. FIDO takes an idea established across domains in machine learning---the ubiquity of scaling laws in data driven systems---and uses it to provide a performance-based stopping criterion. FIDO accomplishes this by acquiring data in small batches, refitting an estimate of the performance curve, and applying a performance-based stopping criterion. FIDO uses a piecewise power law technique grounded in different phases of data collection to produce an accurate estimate of the performance curve. 


Our empirical investigation of FIDO revealed findings with practical implications for the implementation of data minimization. Specifically, certain performance curve families (e.g., the three-parameter power law) systematically overcollect data by not modeling each data collection phase separately. We also demonstrate how active feature acquisition---a technique which might be thought of as a go-to tool for data minimization---can be undesirable in the context of personal data protection. We found that AFA can place excess data collection burden on a small set of users, and that the technique's performance depends on data initialization conditions, with degrading performance in simulations of evolving user communities or restricted feature sets

In light of these complexities, we believe that data minimization compliance in machine learning models will require similar efforts as the principle of fairness has garnered. Definitions, formal implementations and caveats will depend on the application domain, the underlying model (e.g., recommendation, classification). The piecewise power law may not be appropriate for all domains, including user-specific data minimization. Moreover, adversarial approaches are possible: a data processor acting in bad faith may choose a model class that requires a large amount of personal data. Further research is necessary to ensure data minimization occurs despite malicious data collection practices.


%
%